\documentclass[11pt,a4paper]{article}
\usepackage[hyperref]{emnlp2020}
\usepackage{times}
\usepackage{latexsym}

\usepackage{microtype}

\usepackage{graphicx}
\usepackage{amsmath}
\newcommand{\bb}[1]{\textbf{#1}}
\usepackage{url}

\usepackage{hyperref}
\makeatletter
\newcommand{\printfnsymbol}[1]{%
  \textsuperscript{\@fnsymbol{#1}}%
}
\usepackage{microtype}

\aclfinalcopy 

\title{Masked ELMo: An evolution of ELMo towards fully contextual RNN language models}

\author{Gregory Senay\Thanks{Authors made equivalent contributions to this work.}\\
  xBrain / Augustus Intelligence\\Menlo Park, CA, USA \\
  \texttt{gregory.senay@gmail.com}  \And
  Emmanuelle Salin\printfnsymbol{1}\Thanks{Work done during internship at xBrain.} \\
  xBrain\\
  Menlo Park, CA, USA \\
\texttt{emmanuelle.salin@gmail.com}}

\date{03 dec 2020}

\begin{document}
\maketitle
\begin{abstract}
This paper presents Masked ELMo, a new RNN-based model for language model pre-training, evolved from the ELMo language model \cite{ELMo}.
Contrary to ELMo which only uses independent left-to-right and right-to-left contexts,
Masked ELMo learns fully bidirectional word representations.
To achieve this, we use the same Masked language model objective as BERT \cite{BERT}.
Additionally, thanks to optimizations on the LSTM neuron, the integration of mask accumulation and bidirectional truncated backpropagation through time, we have increased the training speed of the model substantially.
All these improvements make it possible to pre-train a better language model than ELMo while maintaining a low computational cost.
We evaluate Masked ELMo by comparing it to ELMo within the same protocol on the GLUE benchmark, where our model outperforms significantly ELMo and is competitive with transformer approaches. 
\end{abstract}

\section{Introduction}
Over the last few years, new language model (LM) pre-training methods have yielded concrete improvements in Natural Language Processing tasks \cite{SeqLearning, ULMFiT, ELMo, BERT, yang2019xlnet}.
These unsupervised methods learn representations of words using a full or partial context.
As a result, the models can extract characteristics of the words and their short and long term relationships, and this in a large variety of contexts.
Lately, Transformer \cite{NIPS2017_7181} approaches such as BERT and XLNet have advanced the state-of-the-art, at the expense of Recurrent Neural Network (RNN) architectures. 
Few studies on RNN-based models have been published since ELMo \cite{ELMo, baevski-etal-2019-cloze}. 
Yet, some limitations of this architecture have been identified in recent papers \cite{BERT} and remain unexplored. 
Indeed, the use of separate left-to-right and right-to-left contexts and short 20-word sentences hurts the pre-training quality. 
Additionally and contrary to Transformer models, ELMo does not clearly benefit from fine-tuning and is most of the time used as a feature-based model (with frozen parameters).

Nevertheless, Transformer-based approaches can be long to train from scratch especially when there is not enough resources available.
Moreover, transformers are usually trained on very large corpus with a long computation time, let uncertainty of their performances when data and corpus are limited.
Maybe in similar conditions, RNN-based approaches can be competitive with transformer.
To reach this goal, the well described limitations of ELMo must be overcome and go beyond. 
We argue that by finding an answer to the shallow bidirectionality of ELMo and by enhancing the training procedure, we can improve the RNN-based pre-trained language models while keeping computational costs at a reasonable level.

In this paper, we present an evolution of ELMo that we call Masked ELMo, which uses the full context when learning the word representations.
This new model, presented in section \ref{Masked ELMo}, is more complex than ELMo in term of recurrent operations and is designed to be use finetuned for downstream tasks.
However, since we improved the computation of the LSTM cell with state projection \cite{HochreiterLSTM} and simplified the rest of the architecture,
we are able to train the model faster while using a larger context (from 20 to 128 tokens).
Section \ref{lm pre-trained} introduces mask accumulation, which ensures a better word coverage during pre-training,
and a proposal for truncated backpropagation through time when done with bidirectional models.
Significant improvements for RNN-based pre-trained LM on different tasks are shown in section \ref{results}.
The paper ends with a conclusion and a discussion about future works.

\section{Masked ELMo}
\label{Masked ELMo}
One of the advantages of RNN-based architectures is that their convergence is faster than Transformer models when only a limited number of resources is available \cite{zhang-etal-2018-accelerating}.
An important step towards reducing the gap between Transformer and RNN language models is to make the recurrent architecture fully contextual.
In fact, a better word representation can be extracted when using the left and right contexts combined, rather than separate \cite{aina-etal-2019-putting}.
Contrary to ELMo, which uses previous and next word predictions independently as pre-training objectives, Masked ELMo uses the same cloze style objective as BERT which predicts masked input tokens.
Since this objective partially masks the information, we can train a bidirectional model with shared left and right contexts without capturing information on the tokens to predict.

Masked ELMo is a 2-layer bidirectional LSTM, and, like ELMo.
It is composed of both left-to-right and right-to-left LSTMs but additional connections are added to make the model fully contextual.
To keep a light architecture in spite of those connections, we do not use Convolutional Neural Network (CNN) over characters to represent the word inputs.
Instead and similarly to \cite{BERT}, we use embedded word-piece representations \cite{WordPiece} to model the input sentences avoiding any out-of-vocabulary in the input and the output of the model.
\begin{figure}[h]
\centering
\includegraphics[width=8cm]{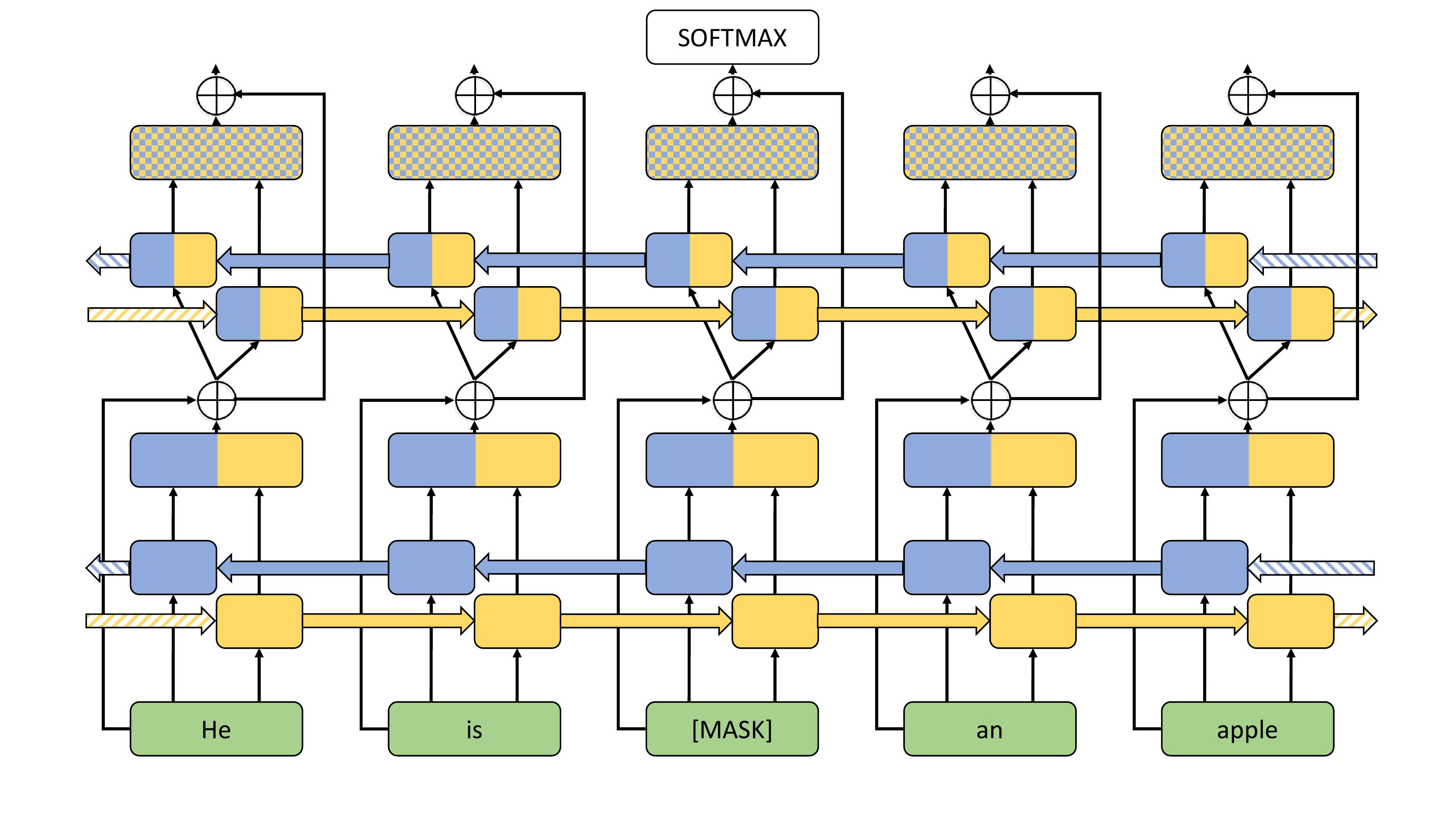} 
\caption{Masked ELMo architecture composed by 2 bidirectional LSTM layers.}
\label{fig:Bidirectional}
\end{figure}
\subsection{Architecture}
The model input is a sequence of N tokens $[t_1,...,t_N]$.
The masked language model computes the probability of this input sequence where $\lfloor Nm \rfloor$ tokens are randomly masked, with $m$ the probability that a token will be masked.
Each masked token $\overline{\rm t_k}$ is predicted given the partially masked context.
\begin{equation*}
  p(t_1, ...,t_N) = \prod_{k=1}^{N} p(t_k|t_1,...,\overline{\rm t_{k}},...,t_{N})
\end{equation*} 

\begin{equation*}
  p(t_1, ...,t_N) = \prod_{k=1}^{N} p(t_1,...,t_{k-1}) p(t_{k+1},...,t_{N})
\end{equation*} 

The input tokens $t_k$ are represented by embeddings of dimension $d$ $[d_1,...,d_k,...,d_N]$ which are followed by $L$ bidirectional LSTM (biLSTM) layers.
Each biLSTM layer $l$ takes an input of dimension $d$ and outputs a vector of dimension $d/2$ for each direction.
These two vectors are then normalized using layer normalization \cite{ba2016layernorm}, and finally, are concatenated into a dimension $d$.
The layer normalization helps to scale vectors across the features, and applying it on the left-to-right and right-to-left outputs helps to improve the performance of the model, since it makes both outputs uniform.
A skip connection is added to all biLSTM layers between the input and the output.
In that case the representation of a token $k$ at layer $l$ is defined as:
\begin{equation*}
  \begin{split}
  h^{LM}_{k,l} = [& LayerNorm(\overrightarrow{h}^{LM}_{k,l}); \\
	              & LayerNorm(\overleftarrow{h}^{LM}_{k,l})] \oplus h^{LM}_{k,l-1}
  \end{split}
\end{equation*}
where $\oplus h^{LM}_{k,l-1}$ is the skip-connection through the biLSTM layer. 
It should be noted that the first biLSTM $l=1$ is not fully contextual as it only takes the embedding representations.
On the contrary, biLSTMs $l>1$ are fully contextual as their inputs contain both left-to-right $\overrightarrow{h}$ and right-to-left $\overleftarrow{h}$ contexts.
The final layer is a softmax aiming at predicting the tokens.
Thanks to these modifications, the model maximises the log likelihood of a token using a truly bidirectional context, which means, it can learn a more complete representation of the words.
Similarly to the ELMo model, the final representation $R_k$ of a token $t_k$ is a set of $L+1$ representations:
\begin{equation*}
  R_{k} = \{h^{LM}_{k,0}, h^{LM}_{k,l} | l=1,...,L\}
\end{equation*}
Furthermore, each representation of a token keeps the same dimension $d$ through all layers, unlike ELMo, which needs to duplicate the first layer output to keep same dimension among layers.

\subsection{LSTM cell with state projection}
In order to improve the computation time of the LSTM cell with state projection, Masked ELMo uses a slightly modified version of the original ELMo LSTM cell.
In this version, all the clippings and the projections are moved outside of the LSTM cell.
This allows us to use an optimized CuDNN LSTM cell \cite{chetlur2014cudnn}.
The only difference between the ELMo LSTM cell and this version is that no clipping is done on the cell state $c_t$ before the computation of the hidden state $h_t$.
This change is illustrated in appendix A, and leads to a 20\% speed-up without increasing the perplexity during pre-training. 
In fact, we notice a slightly lower perplexity. 
Moreover, we think it can be also applied to speed-up ELMo model.
\begin{figure}[ht!]
\centering
\includegraphics[width=6cm]{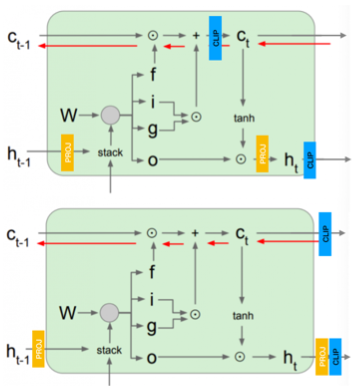} 
\caption{LSTM cell with projection for CUDNN optimization, the cell on the top is the original ELMo version, bellow Masked ELMo modifications.}
\label{fig:LSTMCell}
\end{figure}

\section{Language Modeling pre-training}
\label{lm pre-trained}
Contrary to \cite{ELMo}, Masked ELMo does not use next and previous token prediction pre-training objectives, but a Masked LM (MLM) objective known as cloze task \cite{taylor1953cloze}.
The pre-training protocol and the construction of the input tokens are similar to the BERT language model pre-training \cite{BERT}.
Indeed, there is no more character-level CNN in Masked ELMo but a word-piece vocabulary, which still prevents out-of-vocabulary words.
This reduces significantly the memory footprint and pre-training time.
Thus, batch size can be increased, causing a 300\% speed-up.
Following the author recommendations, a total of 15\% of those tokens are randomly masked: among them 80\% are replaced by a [MASK] token, 10\% are randomly replaced by a different tokens, and the final 10\% are kept as is.
As a result, Masked ELMo takes advantage of this change to increase the context length from 20 words in ELMo to 128 tokens.

\subsection{Mask Accumulation}
Originally, at each epoch of the Masked ELMo pre-training, only 15\% of the input tokens are predicted, compared to 100\% for the ELMo model.
In order to ensure a broader coverage of the tokens, we use a batch accumulation process done on the masked tokens: Mask Accumulation.
During each pre-training step, 4 mutually exclusive sets of tokens are extracted from the input sentence.
Gradients of the same batch masked differently are accumulated and averaged before updating the weights of the model.
For each of those 4 gradient accumulations, the initial states of the LSTM layers are kept the same, to prevent sharing information between those iterations.
During each epoch of the training, a total 60\% of the tokens are masked and predicted, which guarantees that the model covers more data with a similar speed (+6\%).


\begin{table*}[ht!]
\begin{center}
\small
\begin{tabular}{|l|c|c|c|}\hline
	Tasks       	    &Small 	    & Medium	&  Large\\ \hline\hline
Steps btw. validations 	&  100  	&   100	    &  1000\\ \hline
Attention               &     N  	&   N		&  Y\\ \hline
Classifier dropout rate &  0.4   	&  0.2	    &  0.2\\ \hline
Classifier hidden dim	&  128  	&  256	    &  512\\ \hline
Max pool projection dim	&  128    &  256	    &  512\\ \hline
\end{tabular}
\end{center}
\caption{\label{glue-hyperparameters} Hyper-parameters used in the GLUE experiments in the different tasks: Small (RTE, WNLI) Medium (SST-2, CoLA, MRPC) and Large (STS-B, QNLI, MNLI and QQP).}
\end{table*}

\begin{table*}[ht!]
\begin{center}
\centering
\small
\begin{tabular}{|l|ccccccccl|}
\hline
\textbf{Model}	& CoLA  & SST-2 & MRPC  & STS-B	&QQP	&MNLI	&QNLI	&RTE	&Average\\
\hline
ELMo (fz)       & 37.5 & 88.2 & 82.9 & 81.0	& 82.8 & 74.1 & 81.6 & 58.5 & 73.3\\
ELMo (ft)       & 39.3 & 91.4 & 81.4 & 82.4 & 86.4 & 76.7 & 83.6 & 58.1 & 74.9\\
ELMo 5.5B (fz)	& 40.7	& 87.6	& 82.3	& 81.6	& 82.9	& 74.8	& 80.8	& 58.1 	& 73.6 \\
ELMo 5.5B (ft) 	& 38.1	& 89.9	& 80.1	& 82.2	& 86.9	& 77.5	& 83.2	& 57.0	& 74.4 \\
Masked ELMo               & 52.4 & 91.3 & 82.3 & 84.8 & 86.6 & 79.4 & 87.7 & 62.8 & 78.4\\ \hline\hline
BERT* & 44.9 & 91.7 & 84.8 & 86.1 & 88.8 & 81.6 & 89.4 & 64.3 & 78.9\\

\hline
\end{tabular}
\end{center}
\caption{\label{glue-dev} Validation scores of Masked ELMo against baselines (BERT is trained on the same corpus as Masked ELMo) on the GLUE task. }
\end{table*}

\subsection{Bidirectional truncated backpropagation through time}
Masked ELMo is trained using truncated backpropagation through time (TBPTT) \cite{HochreiterLSTM, tbptt} like ELMo and other Language Models \cite{jozefowicz2016exploring} to deal with problems such as a slow training and gradient vanishing.
However, especially in the case of ELMo and Masked ELMo, models can be bidirectional, which partially breaks the TBPTT.
Let $\overrightarrow{s}_n^b$ and $\overleftarrow{s}_n^b$ be the forward and backward states of a LSTM cell of a batch $b \in [0,..,B]$ at position $n \in [0,...,N]$ where
$B$ is the number of batches and $N$ is the sequence length. 
With TBPTT, a batch $b$ is usually initialized with the states $[\overrightarrow{s}_{N}^{b-1};\overleftarrow{s}_{0}^{b-1}]$, providing only information from the past batches. 
Nevertheless, only $\overrightarrow{s}_{N}^{b-1}$ follows the correct word order through batches, unlike $\overleftarrow{s}_{0}^{b-1}$ which ideally should be $\overleftarrow{s}_0^{b+1}$ to correctly initialize $b$.

To counterbalance this, we choose to iterate on the reverse order of batches half of the time, meaning the sequences in a batch are split between the regular order and the reverse order.
We name it Bidirectional TBPTT (BTBPTT).
We argue that this BTBPTT can take advantage of longer dependencies across batches, and could also be used during ELMo training.

\subsection{Unsupervised pre-training}
Masked ELMo pre-training is done on a subset of the Wikipedia English and Book Corpus, with an ablation study realized on the 1-billion corpus dataset \cite{Chelba2013OneBW} (OneBW) used in ELMo. 
The size of our dataset is limited to be similar in size to OneBW, and we introduce a document-level shuffling in order to keep long contiguous sentences rather than just a shuffling at the sentence level like OneBW.
We hope that the model will improve by relying on distant context and longer dependencies.

In addition, the original ratio between Wikipedia and Book Corpus is kept.
A holdout is extracted for validation to compute the masked perplexity. 
The dataset is tokenized using the \textit{bert-base-cased} \cite{Wolf2019HuggingFacesTS}, leading to a total number of 28,745 tokens.
Paragraphs start with \textit{[CLS]} token and sentences are separated with the \textit{[SEP]} token.


Masked ELMo uses 2 biLSTM layers with 4096 units (like ELMo) and 384 dimension projections for each direction, leading to 104M parameters. 
Pre-training is done using the Adam optimizer \cite{adam} with a learning rate of 0.001 and a decay of 0.95.
The model is trained on 10 epochs with 4 mask accumulations which means covering the corpus 40 times. 
The perplexity reached 5.45 on the holdout after 7.8 days of computation with 4*NVIDIA 1080Ti.
ELMo needs 10 days to be trained with the same setup.




\section{Supervised experiments}
\label{results}
For a fair comparison, Masked ELMo is evaluated within the same experimental protocol as ELMo on the GLUE benchmark \cite{wang2018glue}.
Both ELMo and Masked ELMo were trained with the same task-specific architecture: a 2-layer biLSTM with 1024 units, with bi-attention for 2-sentence tasks and a softmax layer.
We keep the parameters proposed in \cite{wang2019jiant} and the scalar-mix introduced in the ELMo paper, table \ref{glue-hyperparameters} shows the used hyper-parameters of the architecture.
We only adjust the Adam optimizer learning rate which is set to 3e-4 for the small tasks (SST-2, CoLA, STS-B, RTE, MRPC, WNLI) and 1e-4 for the larger tasks (QNLI, MNLI and QQP).

Experiments are conducted with and without fine-tuning ELMo. 
It should be noted that ELMo 5.5B is trained on a larger corpus than ELMo and Masked ELMo (Wikipedia: 1.9B and the common crawl from WMT 2008-2012: 3.6B).
Moreover, a BERT baseline (BERT*) trained on the same corpus and a same 128 token context size as Masked ELMo is reported (the training time needed 2 more days in the same setup).


\section{Results}

\begin{table*}[ht]
\begin{center}
\centering
\small
\begin{tabular}{|l|ccccccccl|}
\hline
\textbf{Model}				&CoLA	&SST-2	&MRPC	&STS-B	&QQP	&MNLI	&QNLI	&RTE	&Average 		\\ \hline
ELMo (fz)	                & 34.6	&89.5	&79.5	&73.3	&71.9		&74.9	&80.3	&55.6	&69.40		\\ \hline
ELMo (ft)                   & 32.8	&89.2	&78.7	&75.7	&67.5		&73.3	&80.5	&53.8	&68.51 	\\ \hline
ELMo 5.5B (fz) 	            &35.8	&87.8	&76.8	&76.2	&75.7		&74.2	&81.2	&53.5	&69.59 		\\ \hline
ELMo 5.5B (ft)              &32.4	&90.8	&76.1	&\bb{78.5}	&76.7		&77.1	&83.4	&52.7	&70.31	\\ \hline
Masked ELMo            &\bb{45.2}	&\bb{91.3}	&\bb{79.1}	&76.2	&\bb{77.6}		&\bb{78.7}	&\bb{87.0}	&58.1	&\bb{73.13} \\ \hline\hline

OpenAI GPT \cite{GPTopenAI}	& 45.1	&91.3	&82.3	&80.0	&70.3	&81.6	&85.3	&56.0	&73.02	\\ \hline
BERT*    &11.5	&92.2	&84.1	&80.3	&79.5	&81.3	&88.9	&65.7	&72.10	\\ \hline
\end{tabular}
\end{center}
\caption{\label{glue-test} Test scores of Masked ELMo against ELMo baselines on the GLUE task.}
\end{table*}

Table \ref{glue-dev} compares Masked ELMo on the GLUE benchmark development sets\footnote{WNLI is not included in the average because the majority voting for all methods performs better.} to ELMo models.
As many studies mention, like \cite{ELMo-tune}, fine-tuning ELMo can hurt the performance.
Here, Masked ELMo is directly finetuned on the downstream task.
Better performances are achieved by Masked ELMo, which yields an average improvement of 3.5\%, with significant gains shown on CoLA, STS-B, MNLI, QNLI and RTE.
It is also interesting that the average performances of Masked ELMo are lightly lower than BERT when trained on the same small corpus as Masked ELMo.
On this side, BERT which is trained on the same corpus, outperforms Masked ELMo except on the CoLA task.

Table \ref{glue-dev} reports the online GLUE test set evaluation.
As in the development set, Masked ELMo outperforms the ELMo (fz) model on all tasks by an average of +3.73\%.
A sizable 2.82\% improvement is kept compared to ELMo 5.5B.
Improvements are consistent across the all the GLUE tasks, only STS-B score if lower (-2.3).
Moreover, Masked ELMo is competitive with OpenAI GPT \cite{GPTopenAI}, albeit GPT is also pre-trained on a much larger corpus than Masked ELMo.
Furthermore, when trained on the same dataset, Masked ELMo outperforms BERT but this is mainly due to the poorly CoLA scores where BERT suffers from this low amount of training data and a limited context size, else BERT outperforms Masked ELMo for all other GLUE tasks.


\section{Ablations}

\subsection{ELMo against Masked ELMo on 1Billion corpus}

\begin{table*}[ht]
\begin{center}
\centering
\small
\begin{tabular}{|l|ccccccccl|}
\hline
\textbf{Model}	& CoLA   & SST-2  & MRPC   &STS-B	&QQP	&MNLI	&QNLI	&RTE	&Average\\
\hline
ELMo (OneBW)	        & 37.45 & 88.19 & 82.85 & 80.97 & 82.78 & 74.06	& 81.57	& 58.48	& 73.295\\
Masked ELMo (OneBW)	& 49.01 & 91.17	& 83.17 & 83.93	& 86.68 & 78.11 & 85.54	& 60.29 & 77.239\\\hline
\end{tabular}
\end{center}
\caption{\label{1b-ablation} Validation scores of Masked ELMo against ELMo baselines trained done the same corpus (1 Billion word).} 
\end{table*}
Table \ref{1b-ablation} reports the impact of Masked ELMo again ELMo when trained on the same dataset: 1-billion word corpus (OneBW).
Results show that the Masked ELMo architecture is able to take a better advantage on the same corpus, outperforming ELMo in all tasks.
This confirms our hypothesis that a fully bidirectional context in RNNs allows to learn a better contextual representation of the words and the sentences.

\begin{table*}[ht!]
\begin{center}
\small
\begin{tabular}{|c|c|c|c|c|c|c|} \hline
seq len & Train & Dev  	& Test	& batch size & words per minibatch\\ \hline\hline
 20     & 18.19 & 21.20 & 20.17	& 128        & 2,560 \\\hline
 20     & 19.16 & 21.08 & 20.53	&  20        &   400 \\\hline
128     & 11.44 & 14.78 & 14.12	&  20        & 2,560 \\\hline
\end{tabular}
\end{center}
\caption{\label{wiki2-seqlen} Impact of the sequence length for Masked ELMo on Wiki-2 corpus.}
\end{table*}
\begin{table*}[ht!]
\begin{center}
\small
\begin{tabular}{|c|c|c|c|} \hline
Wiki-2           	&           Train         	&          Dev		&        Test		\\ \hline\hline
w/o bd-tbptt 	&  21.53 $\pm$ 6.25	& 22.00 $\pm$ 0.95	& 21.08 $\pm$ 0.62	\\ \hline
bd-tbptt         	&  18.96 $\pm$ 0.50	& 21.33 $\pm$ 0.23	& 20.64 $\pm$ 0.11 	\\ \hline
\end{tabular}
\end{center}
\caption{\label{wiki2-bdtbptt} Impact of the bidirectional truncated back propagation through time on Wiki-2 corpus.}
\end{table*}
\subsection{Impact of the sequence length in training}

Additionally, we have conducted ablation studies on the Wiki-2 corpus.
Table \ref{wiki2-seqlen} shows the perplexity of Masked ELMo when using a 20 and 128 sequence length.
For this model is trained with a lighter version of the Masked ELMo architecture (\#31M parameters with layer projection size of 512, internal state size before projection 2048).
The model is trained on 10 epochs with an initial learning rate of 0.001.

As results shown, the impact of the sequence length is important for reducing the perplexity.

\subsection{Impact of BD-TBPTT}
Last ablation study, reports the effect of the bidirectional truncated back-propagation through time.
Table \ref{wiki2-bdtbptt} presents an experiments on Wiki-2 corpus using a batch size of 20 and a sequence length of 20 tokens which forces the model to rely more on the RNN initial states compared to a longer sequence of 128, where most of the context is already available in the sequence.
This is interesting to noticed that the bidirectional TBPTT leads to a small but consistent improvements on the perplexity, and more interesting a much lower standard deviation over multiple runs.

\section{Conclusion and perspective}
We present Masked ELMo, an evolution of ELMo, which takes advantage of joint left-to-right and right-to-left contexts using a fully bidirectional architecture.
We also introduce mask accumulation, which allows better data coverage, and BTBPTT, an other way to do the TBPTT for bidirectional architectures.
This model outperforms ELMo in the GLUE benchmark while being more computationally efficient.
As results show, it could be interesting to train Masked ELMo on a larger dataset.
Moreover, additional biLSTM layers would provide a deeper contextual representation.
Finally, averaging the biLSTM forward and backward outputs, instead of concatenating them, would project these features in the same space.
We hope this model is a step forward to improve RNN architectures for pre-trained language model.




\bibliographystyle{acl_natbib}
\bibliography{masked_elmo}

\end{document}